\documentstyle[chicagob,12pt]{article}

\newtheorem{THEOREM}{Theorem}[section]
\newenvironment{theorem}{\begin{THEOREM} \hspace{-.85em} {\bf :} }%
                        {\end{THEOREM}}
\newtheorem{LEMMA}[THEOREM]{Lemma}
\newenvironment{lemma}{\begin{LEMMA} \hspace{-.85em} {\bf :} }%
                      {\end{LEMMA}}
\newtheorem{COROLLARY}[THEOREM]{Corollary}
\newenvironment{corollary}{\begin{COROLLARY} \hspace{-.85em} {\bf :} }%
                          {\end{COROLLARY}}
\newtheorem{PROPOSITION}[THEOREM]{Proposition}
\newenvironment{proposition}{\begin{PROPOSITION} \hspace{-.85em} {\bf :} }%
                            {\end{PROPOSITION}}
\newtheorem{DEFINITION}[THEOREM]{Definition}
\newenvironment{definition}{\begin{DEFINITION} \hspace{-.85em} {\bf :} \rm}%
                            {\end{DEFINITION}}
\newtheorem{CLAIM}[THEOREM]{Claim}
\newenvironment{claim}{\begin{CLAIM} \hspace{-.85em} {\bf :} \rm}%
                            {\end{CLAIM}}
\newtheorem{EXAMPLE}[THEOREM]{Example}
\newenvironment{example}{\begin{EXAMPLE} \hspace{-.85em} {\bf :} \rm}%
                            {\end{EXAMPLE}}
\newtheorem{REMARK}[THEOREM]{Remark}
\newenvironment{remark}{\begin{REMARK} \hspace{-.85em} {\bf :} \rm}%
                            {\end{REMARK}}

\newcommand{\thm}{\begin{theorem}}
\newcommand{\lem}{\begin{lemma}}
\newcommand{\pro}{\begin{proposition}}
\newcommand{\dfn}{\begin{definition}}
\newcommand{\rem}{\begin{remark}}
\newcommand{\xam}{\begin{example}}
\newcommand{\cor}{\begin{corollary}}

\newcommand{\ethm}{\end{theorem}}
\newcommand{\elem}{\end{lemma}}
\newcommand{\epro}{\end{proposition}}
\newcommand{\edfn}{\bbox\end{definition}}
\newcommand{\erem}{\bbox\end{remark}}
\newcommand{\exam}{\bbox\end{example}}
\newcommand{\ecor}{\end{corollary}}

\newcommand{\beqn}{\begin{equation}}
\newcommand{\eeqn}{\end{equation}}

\newcommand{\bbox}{\vrule height7pt width4pt depth1pt}

\newcommand{\clm}{\begin{claim}}
\newcommand{\eclm}{\end{claim}}

\newcommand{\sat}{\models}

\newcommand{\rimp}{\Rightarrow}

\newcommand{\dimp}{\Leftrightarrow}
\newcommand{\bor}{\bigvee}

\newcommand{\union}{\cup}

\renewcommand{\phi}{\varphi}

\newcommand{\cross}{\times}

\newcommand{\R}{{\cal R}}

\newcommand{\<}{\langle}
\renewcommand{\>}{\rangle}

\newcommand{\ie}{i.e.,~}

\newcommand{\cf}{cf.~}

\newcommand{\ol}{\setlength{\itemsep}{0pt}\begin{enumerate}}
\newcommand{\eol}{\end{enumerate}\setlength{\itemsep}{-\parsep}}
\newcommand{\ul}{\setlength{\itemsep}{0pt}\begin{itemize}}
\newcommand{\dl}{\setlength{\itemsep}{0pt}\begin{description}}
\newcommand{\edl}{\end{description}\setlength{\itemsep}{-\parsep}}
\newcommand{\eul}{\end{itemize}\setlength{\itemsep}{-\parsep}}

\newcommand{\false}{\mbox{{\it false}}}

\newcommand{\commentout}[1]{}

\newcommand{\bi}{\begin{itemize}}
\newcommand{\ei}{\end{itemize}}
\newcommand{\be}{\begin{enumerate}}
\newcommand{\ee}{\end{enumerate}}

\newcommand{\denselist}{\itemsep 0pt\partopsep 0pt}

\renewcommand{\L}{{\cal L}}

\newcommand{\RCond}{>}

\newcommand{\Bel}{B}

\newcommand{\False}{\mbox{\it false}}

\newcommand{\Cref}[1]{Corollary~\ref{#1}}

\newcommand{\BEL}{\mbox{Bel}}

\newenvironment{RETHM}[2]{\it \trivlist \item[\hskip \labelsep{\bf #1 \ref{#2}:}]}{\endtrivlist}
\newcommand{\rethm}[1]{\begin{RETHM}{Theorem}{#1}}
\newcommand{\repro}[1]{\begin{RETHM}{Proposition}{#1}}
\newcommand{\relem}[1]{\begin{RETHM}{Lemma}{#1}}
\newcommand{\recor}[1]{\begin{RETHM}{Corollary}{#1}}

\newcommand{\erethm}{\end{RETHM}}
\newcommand{\erepro}{\end{RETHM}}
\newcommand{\erelem}{\end{RETHM}}
\newcommand{\erecor}{\end{RETHM}}

\newenvironment{oldthm}[1]{\par\noindent{\bf Theorem #1:} \em \noindent}{\par}
\newenvironment{oldlem}[1]{\par\noindent{\bf Lemma #1:} \em \noindent}{\par}
\newenvironment{oldcor}[1]{\par\noindent{\bf Corollary #1:} \em \noindent}{\par}
\newenvironment{oldpro}[1]{\par\noindent{\bf Proposition #1:} \em \noindent}{\par}
\newcommand{\othm}[1]{\begin{oldthm}{\ref{#1}}}
\newcommand{\eothm}{\end{oldthm} \medskip}
\newcommand{\olem}[1]{\begin{oldlem}{\ref{#1}}}
\newcommand{\eolem}{\end{oldlem} \medskip}
\newcommand{\ocor}[1]{\begin{oldcor}{\ref{#1}}}
\newcommand{\eocor}{\end{oldcor} \medskip}
\newcommand{\opro}[1]{\begin{oldpro}{\ref{#1}}}
\newcommand{\eopro}{\end{oldpro} \medskip}

\newcommand{\bxor}[1]{\dot{\bor}}

\newcommand{\RevDP}{\Rev_{DP\,}}
\newcommand{\RevB}{\Rev_{B\,}}

\newcommand{\Rev}{*}
\newcommand{\Cl}{\mbox{Cl}}

\renewcommand{\Bel}{\mbox{\it Bel\/}}

\begin{document}
\begin{titlepage}
\title{Belief Revision: A Critique%
\thanks{
Some of this work was done while both authors were at the IBM Almaden
Research Center. The first author was also at Stanford while much of
the work was done.  IBM and Stanford's support are gratefully
acknowledged.  This work was also supported in part by NSF
under grants IRI-95-03109 and IRI-96-25901, by the Air Force Office of
Scientific Research under grant F49620-96-1-0323, and
by an IBM Graduate Fellowship
to the first author.
A preliminary version of this paper appeared in
L.~C. Aiello, J.~Doyle, and S.~C. Shapiro (Eds.)
{\em Principles of knowledge representation and reasoning :
Proc.~Fifth International Conference (KR '96)}, pp.~421--431, 1996.
This version will appear in the {\em Journal of Logic, Language, and
Information}. 
}
}
\author{Nir Friedman\\
\small Institute of Computer Science\\
\small Hebrew University \\
\small Jerusalem, 91904, Israel\\
\small nir@cs.huji.ac.il\\
\small http://www.cs.huji.ac.il/$\sim$nir
\and
Joseph Y.\ Halpern\\
\small Computer Science Department\\
\small Cornell University\\
\small Ithaca NY 14853.\\
\small halpern@cs.cornell.edu\\
\small http://www.cs.cornell.edu/home/halpern
}
\date{\today}
\setcounter{page}{0}
\maketitle
\thispagestyle{empty}
\begin{abstract}
We examine carefully the rationale underlying
    the approaches to belief change  taken in the literature,
and highlight what we view as methodological problems.
We argue that to study belief change carefully, we must
be quite explicit about the ``ontology'' or scenario
underlying the belief change process.  This is something that has been
missing in previous work, with its focus on postulates.
Our analysis shows that we must pay particular attention to two
    issues that have often been taken for granted:
The first is how we model the agent's epistemic state.  (Do we use a
set of beliefs, or a
richer structure, such as an ordering on worlds? And if we use a set of
beliefs, in what
language are these beliefs are expressed?)
We show that even postulates that have been called ``beyond
controversy'' are
    unreasonable when the agent's beliefs include beliefs about
    her own epistemic state as well as the external world.
The second is the status of
observations.  (Are observations known to be true, or just believed?  In
the latter case, how firm is the belief?)
Issues regarding
 the status of observations arise particularly when we consider
{\em iterated\/} belief revision, and we must confront the possibility
of revising by $\phi$ and then by $\neg \phi$.
\end{abstract}

\end{titlepage}

\section{Introduction}

The problem of {\em belief change\/}---how an agent should revise her
beliefs upon learning new information---has been an active area of
research in both philosophy and artificial intelligence.
The problem is a fascinating one in part
because it is clearly no unique answer.  Nevertheless, there
is a strong intuition that one wants to make {\em minimal\/} changes,
and all the approaches to belief change in the literature, such as
\cite{agm:85,Gardenfors1,KM91}, try to incorporate this principle.
However, approaches differ on what constitutes a minimal change.
This issue has come to the fore with the spate of recent work on
{\em iterated\/} belief revision (see, for example,
\cite{Boutilier:Iterated,BoutilierGoldszmidt,Darwiche94,FreundLehmann,%
Lehmann95,Levi88,Williams94}).

The approaches to belief change
typically start with a collection of postulates, argue
that they are reasonable, and prove some consequences of these
postulates.
Often
a semantic model for the postulates is provided and a
representation theorem is proved (of the form that every semantic
model corresponds to some belief revision process, and that every belief
revision process can be captured by some semantic model).
Our goal
here
is not to introduce yet another model of
belief change, but
to examine carefully the rationale underlying the
approaches in the literature.
The main message of the paper is that describing postulates and proving
a representation theorem is not enough.
While it may have been reasonable when research on belief change
started in the early 1980s to just consider the implications of a number
of seemingly reasonable postulates, it is our view that it
should no longer be acceptable
just to write down postulates and give
short English justifications for them.
While postulates do provide insight and guidance,
it is also important to
describe what, for want of a better word, we call the underlying
{\em
ontology\/} or scenario for the belief change process. Roughly speaking,
this means
describing carefully what it means for something to be believed by an
agent and what the status is of new information that is received by the
agent.  This point will hopefully become clearer as we present our
critique.  We remark that even though the issue of ontology is tacitly
acknowledged
in a number of papers (for example, in the last paragraph of
\cite{Lehmann95}), it rarely enters into the discussion in a significant
way.%
\footnote{A recent manuscript
by Hansson \citeyear{Hansson98b} does raise some issues of
ontology in the context of justification of beliefs, but no particular
ontology for belief revision is discussed.}
We hope to show that
 ontology must play a central role in all discussions of belief
revision.

Our focus is on approaches that take as their starting point the
postulates for belief revision proposed by Alchourr\'on, G\"ardenfors,
and Makinson (AGM from now on) \citeyear{agm:85}, but our critique
certainly applies to other approaches as well,
in particular, Katsuno and Mendelzon's {\em belief update\/}  \citeyear{KM92};
see Section~\ref{conclusion}.
The AGM approach assumes that an agent's epistemic state is represented
by a {\em belief set}, that is, a set $K$ of formulas in a logical
language $\L$.%
\footnote{For example, G\"ardenfors~\citeyear[p.~21]{Gardenfors1} says
``A simple way of modeling the epistemic state of an individual is to
represent it by a {\em set} of sentences.''}
What the agent learns is assumed to be characterized by some formula
$\phi$, also in $\L$;  $K * \phi$ describes
the belief set of an agent that starts with belief set $K$ and learns
$\phi$.%
\footnote{One of the reviewers considered
the usage of the terms
{\em learns\/} and {\em observes\/} as introducing a
bias regarding the nature of the belief revision process.
We feel that this comment highlights the need for a more
concrete ontology for belief revision.
We continue to use these terms, since they are quite standard in the
literature, but we admit that they are indeed biased.
There may well be ontologies for which they are inappropriate.}

There are two assumptions implicit in this notation:

\begin{itemize}\denselist
\item The functional form of $*$ suggests that all that matters
regarding how an
agent revises her beliefs is the belief set and what is learnt.

\item
The notation suggests that the second argument of $*$ can be an
arbitrary formula
in $\L$.  But what does it mean to revise by $\false$?  In what sense
can $\false$ be learnt?  More generally, is it reasonable to assume
that an arbitrary formula can be learnt in a given epistemic state?
\end{itemize}

The first assumption is particularly problematic when we
consider the postulates that AGM require $\Rev$ to satisfy. These
essentially state that the agent is consistent in her choices,
in the sense that she
acts as though she has an ordering on the strength of her beliefs
\cite{MakGar,Grove}, or an ordering on possible worlds
\cite{Boutilier94AIJ2,Grove,KM92}, or some other
predetermined manner
of choosing among competing beliefs \cite{agm:85}.
However,
the fact that an agent's epistemic state is characterized by a
collection of formulas means that the epistemic state cannot include
information about relative strength of beliefs (as required for the
approach of, say, \cite{MakGar}), unless this information is expressible
in the language.  Note that if $\L$ is propositional logic or
first-order logic, such information cannot be expressed.  On the other
hand, if $\L$ contains {\em conditional\/} formulas
of the form $p \RCond q$, interpreted as ``if $p$ is learnt, then $q$
will be believed'',
then constraints on the relative strength of
beliefs can be expressed (indirectly, by describing which beliefs
will be retained after a revision).

Problems arise when the language is not rich enough to
express relative degrees of strength in beliefs.  Consider, for
example, a situation where $K = Cl(p \land q)$ (the logical closure of
$p \land q$; that is, the agent's beliefs are characterized by the
formula $p \land q$ and its logical consequences), and then the agent
learns $\phi = \neg p \lor \neg q$.  We can imagine that an agent
whose belief in $p$ is stronger than her belief in $q$ would have $K *
\phi = Cl(p)$.  That is, the agent gives up her belief in $q$, but
retains a belief in $p$.  On the other hand, if the agent's belief in
$q$ is stronger than her belief in $p$, it seems reasonable to expect
that $K * \phi = Cl(q)$.  This suggests that it is unreasonable to
take $*$ to be a function if the representation language is not rich
enough to express what may be significant details of an agent's
epistemic state.

We could, of course, assume that information about the relative strength
of beliefs in various propositions is implicit in the choice of the
revision operator $*$, even if it is not contained in the language.
This is perfectly reasonable, and also makes it more reasonable that $*$
be a function.  However, note that we can then no longer assume that we
use the same $*$ when doing iterated revision, since there is no reason
to believe that the relative strength of beliefs is maintained after we
learn a formula.  In fact, in a number of recent papers
\cite{Boutilier:Iterated,BoutilierGoldszmidt,FrH2Full,Nayak:1994,spohn:88,Williams94},
$*$ is
defined as a
function from (epistemic states $\times$ formulas) to epistemic states,
but the epistemic states are no longer just belief sets; they include
information regarding relative strengths of beliefs.
The revision function on epistemic states induces a mapping from (belief
sets $\times$ formulas)
to belief sets, but at the level of belief sets, the mapping may
not be functional; for a belief set $K$ and formula $\phi$, the
belief set $K \Rev\phi$ may depend on what epistemic state induced $K$.
Thus, the effect of $*$ on belief sets may change over time.%
\footnote{Freund and Lehmann
\citeyear{FreundLehmann} have called the viewpoint that $*$ may
change over time the {\em dynamic\/} point of view.  However, this seems
somewhat of a misnomer when applied to papers such as
\cite{Boutilier:Iterated,BoutilierGoldszmidt,FrH2Full,Williams94},
since there $*$ in
fact is static, when viewed as a function on epistemic states and
formulas.}

There is certainly no agreement on what postulates belief change should
satisfy.  However, the following two postulates are almost universal:
\begin{itemize}\denselist
\item $\phi \in K*\phi$
\item if $K$ is consistent
and $\phi \in K$,
then $K * \phi = K$.
\end{itemize}
These postulates have been characterized by Rott \citeyear{Rott} as
being ``beyond controversy''.  Nevertheless, we argue that they are not
as innocent as they may at first appear.

The first postulate says that the agent believes the last thing she
learns.  Making sense of this requires some discussion of the underlying
ontology.  It certainly makes sense if we take (as G\"ardenfors
\citeyear{Gardenfors1} does) the belief set $K$ to consist of all
formulas that are accepted by the agent (where ``accepted'' means
``treated as true''), and the agent revises by $\phi$ only if
$\phi$ has somehow come to be accepted.  However, note that
deciding when a formula has come to be accepted is nontrivial.
In particular, just observing $\phi$ will not in general be
enough for accepting $\phi$.  Acceptance has a complex interaction
with what is already believed.  For example, imagine a scientist who
believes that heavy objects drop faster than light ones, climbs the
tower of Pisa, drops a 5 kilogram textbook and a 500 milligram novel,
and observes they hit the ground at the same time.
This scientist will probably not accept that the time for an object to
fall to the ground is independent of its weight, on the basis of this
one experiment (although perhaps repeated experiments may lead her
to accept it).

While the acceptance point of view is certainly not unreasonable,
the fact that
just observing $\phi$ is not necessarily enough for
acceptance often seems forgotten.
It also seems hard to believe that $\false$ would ever be accepted.
More generally, it is far from obvious that in a given epistemic state
$K$ we should allow arbitrary consistent formulas to be
accepted.  Intuitively, this does not allow for the possibility
that some beliefs are held so firmly that their negations could never
be accepted.  (Later we describe an ontology where observations are
taken to be known in which in fact some consistent formulas will
not be accepted in some epistemic states.)

While the second postulate is perhaps plausible if we cannot talk
about epistemic importance or strength of belief in the language,
it is less so once we can talk about such things (or if either epistemic
importance or strength of
belief is encoded in the epistemic state some other way).
For suppose that $\phi \in K$.
Why should $K * \phi = K$?
It could well be that being informed of $\phi$ raises the importance
of $\phi$ in the epistemic ordering, or the agent's strength of belief
in $\phi$.  If strength of belief can be talked
about in the language, then a notion of minimal change
should still allow strengths of belief to change, even when something
expected is observed.
Even if we cannot talk about strength of belief in the language, this
observation has
an impact on iterated revisions.  For example, one assumption made by
Lehmann \citeyear{Lehmann95} (his postulate I4) is that if $p$ is
believed after revising by $\phi$, then revising by $[\phi \cdot p \cdot
\psi]$---that is, revising by $\phi$ then $p$ then $\psi$---is
equivalent to revising by $[\phi \cdot \psi]$.
But consider a situation where after revising by $\phi$, the agent
believes both $p$ and $q$, but her belief in $q$ is stronger than her
belief in $p$.  We can well imagine that after learning $\neg p \lor
\neg q$ in this situation, she would believe $\neg p$ and $q$.  However,
if she first learned $p$ and then $\neg p \lor \neg q$, she would believe
$p$ and $\neg q$, because, as a result of learning $p$, she would give
$p$ higher epistemic importance than $q$.
In this case, we would not have
$[\phi \cdot p \cdot (\neg p \lor \neg q) ] =
[\phi \cdot (\neg p \lor \neg q)]$.
In light of this discussion,
it is not surprising that the combination
of the second postulate with a language that can talk about epistemic
ordering  leads to technical problems such as G\"ardenfors' {\em
triviality result\/} \citeyear{Gardenfors1}.

\commentout{
This discussion already shows why
G\"ardenfors' celebrated {\em triviality result\/}
\citeyear{Gardenfors1} arises.
This result essentially says that the AGM postulates
cannot be applied to belief sets that contain {\em Ramsey
conditionals\/} of the form $\phi \RCond \psi$ with the interpretation
``revising by $\phi$ will lead to a state where $\psi$ is believed''.
With such formulas in the belief sets, we can see which formulas will
be discarded when something inconsistent is learnt, so we effectively
get an ordering of epistemic importance.  As we have observed, the
second postulate (which is one of the AGM postulates) is not sound in
this case.  The triviality result should thus come as no surprise.
(This point is essentially also made in \cite{FrH2Full,Wobcke95}.)
}

To give a sense of our concerns here, we discuss two basic ontologies.
The first ontology that seems (to us) reasonable assumes that the
agent has some knowledge as well as beliefs.
We can think of the formulas that the agent knows as having
the highest state of epistemic importance.  In keeping with the standard
interpretation of knowledge, we also assume that the formulas that the
agent knows are true in the world.  Since agents typically do not have
certain knowledge of very many facts, we assume that the knowledge
is augmented by beliefs (which can be thought of as defeasible guides to
action).  Thus, the set of formulas that are known form a subset of the
belief set.   We assume that the agent
observes the world using
reliable sensors; thus, if the agent observes $\phi$, then the agent is
assumed to know $\phi$.%
\footnote{In fact, the phrase ``reliable sensors'' is
somewhat too strong. We can deal with unreliable sensors in our
framework by explicitly modeling the difference between the sensor
reading the agent observes, which we take to be a reliable
observation, and the actual state of the external world that the
sensor is, unreliably, measuring or reporting. See \cite{BFH1} for
a more
detailed
discussion of this point.}
After observing $\phi$, the agent adds $\phi$
to his stock of knowledge, and may revise his belief set.
Since the agent's observations are taken to be knowledge, the
agent will believe $\phi$ after observing $\phi$.
However,
the agent's epistemic state may change even if she observes a
formula
that she previously believed to be true.  In particular, if the formula
observed was believed to be true but not known to be true, after the
observation it is known.
Note that, in this ontology,
the agent never observes $\false$, since $\false$ is not true of
the world.  In fact, the agent never observes anything that contradicts
her knowledge.  Thus, $K * \phi$ is defined only for formulas $\phi$
that are compatible with the agent's knowledge.
Moving to iterated revision, this means we cannot have a revision by
$\phi$ followed by a revision by $\neg \phi$.
This ontology underlies some of our earlier work
\citeyear{FrThesis,FrH1Full,FrH2Full}.
As we show here, by taking a variant of Darwiche
and Pearl's approach \citeyear{Darwiche94}, we can also capture this
ontology.
This ontology captures the essence of Bayesian updating in probabilistic
reasoning.  By taking observations to be known, we are essentially
giving observed events probability 1 and conditioning on them.

\commentout{
The first ontology presumes that what is observed is known to be true.
This idea can be generalized:  We can associate with each observation
the strength of belief it should have in the new epistemic state.
If what is observed is known to be true, then it has the strongest
possible strength of belief.  But we can stipulate that it should have a
weaker strength of belief.  Darwiche and Pearl \citeyear{Darwiche94}
actually take it to have the weakest possible strength of belief.
Following the lead of Spohn \citeyear{spohn:88},
Goldszmidt and Pearl \citeyear{Goldszmidt92},
Williams
\citeyear{Williams94} and Wobcke \citeyear{Wobcke95} allow generalized
observations, that include the strength of belief they should have
in the new epistemic state.}

We can consider a second ontology that has a different flavor.
In this ontology,
if we observe something, we believe it to be true and perhaps even
assign it a strength of belief.  But this assignment does not represent
the strength of belief of the observation in the resulting
epistemic state. Rather, the belief in the observation  must ``compete''
against current beliefs if it is inconsistent with these beliefs.
In this ontology, it is not necessarily the case
that $\phi \in K * \phi$, just as it is not the case that
a scientist will necessarily adopt the consequences of his most recent
observation into his stock of beliefs (at least, not without
doing some additional experimentation).
Of course, to flesh out this ontology, we need to describe how to
combine a given strength of belief
in the observation with the strengths of the beliefs in the original
epistemic state.
Perhaps the closest parallel in the
uncertainty
literature is something like the
Dempster-Shafer rule of combination \cite{Shaf}, which gives a rule for
combining two separate bodies of belief.
We believe that this
type of ontology deserves further study.
We sketch one particular approach to modeling this ontology in
Section~\ref{conclusion}
and refer the reader to \cite{BFH1} for a
more thorough treatment of
it.
A quite different treatment of this problem, more in the spirit of
the AGM approach, has been studied by Hansson \citeyear{Hansson91},
Makinson \citeyear{Makinson97}, and
others; see \cite{Hansson98a} for an overview of this work.

The rest of the paper is organized as follows.  In Section~\ref{AGM}, we
review the AGM framework, and point out some problems with it.
In Section~\ref{otherproposals}, we consider proposals for
belief change and iterated belief change from the
literature due to Boutilier \citeyear{Boutilier:Iterated}, Darwiche and Pearl
\citeyear{Darwiche94}, Freund and Lehmann \citeyear{FreundLehmann}, and
Lehmann \citeyear{Lehmann95}, and try
to understand the ontology implicit in the proposal (to the extent
that one can be discerned).  In Section~\ref{observationknowledge},
we consider the first ontology discussed above
in more detail.  We conclude with some
discussion in Section~\ref{conclusion}.

\section{AGM Belief Revision}\label{AGM}
In this section we review the AGM approach to belief revision.
As we said earlier, this approach assumes that beliefs and observations
are expressed in some language $\L$.  It is assumed that $\L$ is closed
under negation and conjunction, and comes equipped with a
compact
consequence relation $\vdash_\L$ that contains
the propositional calculus and satisfies the deduction theorem.  The
agent's epistemic state is represented by a belief set, that
is, a set of formulas in $\L$ closed under deduction.  There is also
assumed to be a revision operator $\Rev$ that takes a belief set $K$
and a formula $\phi$ and returns a new belief set $K \Rev \phi$,
intuitively, the result of revising $K$ by $\phi$. The following AGM
postulates are an attempt to characterize the intuition of ``minimal
change'':
\begin{description}\denselist
 \item[R1.] $K \Rev \phi$ is a belief set
 \item[R2.] $\phi \in K\Rev\phi$
 \item[R3.] $K \Rev\phi \subseteq Cl(K \cup \{\phi\})$%
 \item[R4.] If $\neg\phi \not\in K$ then $Cl(K \cup \{\phi\})
    \subseteq K \Rev\phi$
 \item[R5.] $K\Rev\phi = Cl(\False)$ if and only if
$\vdash_\L \neg\phi$
 \item[R6.] If
$\vdash_\L \phi \dimp \psi$
then $K\Rev\phi = K\Rev\psi$
 \item[R7.] $K \Rev (\phi\land\psi) \subseteq Cl(K\Rev\phi \cup \{\psi\})$
 \item[R8.] If $\neg\psi \not\in K\Rev\phi$ then $Cl(K\Rev\phi \cup
    \{\psi\}) \subseteq K \Rev (\phi\land\psi)$
\end{description}

The essence of these postulates is the following.  Revising $K$ by
$\phi$ gives a belief set (Postulate R1) that includes $\phi$ (R2).  If
$\phi$ is consistent with $K$, then $K \Rev \phi$
consist precisely of those beliefs
implied by the combination of the old beliefs
with the new belief (R3 and R4).
Note that it follows from R1--R4 that if $\phi \in K$, then
$K = K \Rev \phi$.
The next two postulates discuss the coherence
of beliefs.  R5 states that as long as $\phi$ is consistent,
then so is $K \Rev \phi$, and R6 states that the
syntactic form of the new belief does not affect the revision process.
The last two postulates enforce a certain coherency on the outcome of
revisions by related beliefs.  Basically they state that if $\psi$ is
consistent with $K\Rev\phi$ then $K\Rev(\phi\land\psi)$ is the result of
adding $\psi$ to $K\Rev\phi$.

The intuitions described by AGM is based on one-step (noniterated)
revision. Nevertheless, the AGM postulates do impose some
restrictions on iterated revisions.  For example, suppose that $q$ is
consistent with
$K \Rev p$. Then, according to R2 and R3, $(K * p) * q = \Cl(K*p \union
\{ q\})$.  Using R7 and R8 we can conclude that $(K * p) * q = K * (p
\land q)$.

There are several representation theorems for AGM belief revision;
perhaps the easiest to understand is due to Grove \citeyear{Grove}.
We discuss a slight modification, due to Boutilier
\citeyear{Boutilier94AIJ2} and Katsuno and Mendelzon \citeyear{KM92}:
Let an {\em ${\cal L}$-world\/} be a complete and
consistent truth
assignment to the formulas in ${\cal L}$.  Let ${\cal W}$ consist of
all the ${\cal L}$-worlds, and let $\preceq$ be a {\em ranking}, that
is, a total preorder, on the worlds in ${\cal W}$.
Let $\min_\preceq$ consist of all
the minimal worlds with respect to $\preceq$, that is, all the worlds
$w$ such that there is no $w'$ with $w' \prec w$.  With $\preceq$ we
can associate a
belief set $\BEL(\preceq)$, consisting of all formulas $\phi$ that are true
in all the worlds in $\min_\preceq$.  Moreover, we can define a
revision operator $\Rev$, by taking $\BEL(\preceq) \Rev
\phi$ to consist of all formulas $\psi$ that are true in all the minimal
$\phi$-worlds according to $\preceq$.  It can be shown
that $\Rev$ satisfies the AGM postulates (when its first argument is
$\BEL(\preceq)$).
Thus,
we can define a revision operator by taking a collection of orderings
$\preceq_K$, one for each belief set $K$.
To define $K \Rev \phi$ for a belief set $K$, we
apply the procedure above, starting with the ranking $\preceq_{K}$
corresponding to $K$.%
\footnote{In this construction, for each belief set $K$ other than
the inconsistent belief set, we have $\BEL(\preceq_K) = K$.  The
inconsistent belief set gets special treatment here.}
Furthermore, Grove \citeyear{Grove}, Katsuno and Mendelzon
\citeyear{KM92}, and Boutilier \citeyear{Boutilier94AIJ2} show
that every belief
revision operator satisfying the AGM axioms can be characterized in
this way.

This elegant representation theorem also brings out some of the problems
with the AGM postulates.
First, note that a given revision operator $\Rev$ is represented by a
family of rankings, one for each belief set.  There is no
necessary connection between the rankings corresponding to different
belief sets.  It might seem more reasonable to have a more global
setting (perhaps one global ranking) from which each element in the
family of rankings arises.

A second important point is that the epistemic state here is a
function not only of the  belief set, but also of the ranking. The
latter, however, is represented only implicitly.
Each ranking $\preceq$ is associated with
a belief set $\BEL(\preceq)$, but it is the ranking
that gives the information required to describe
how revision is carried out.  The belief set does not suffice to
determine the revision; there are many rankings $\preceq$ for
which the associated belief set $\BEL(\preceq)$ is $K$.
Since the revision process only gives us the revised belief set, not the
revised ranking, the representation does not support iterated
revision.

This suggests that we should consider, not how to revise
belief sets, but how to revise rankings.  More generally,
whatever we take to be
our representation of the epistemic state, it seems appropriate to
consider how these representations should be revised.
We can define an analogue of the AGM postulates for epistemic states
in a straightforward way
(\cf \cite{FrH2}):
Taking $E$ to range over epistemic
states and $\Bel(E)$ to represent the belief set associated with
epistemic state $E$, we have
\begin{description}\denselist
 \item[R1$'.$] $E \Rev \phi$ is an epistemic state
 \item[R2$'.$] $\phi \in \Bel(E\Rev\phi)$
 \item[R3$'.$] $\Bel(E \Rev\phi) \subseteq Cl(\Bel(E) \cup \{\phi\})$
\end{description}
and so on, with the obvious syntactic transformation.
In fact, as we shall see in the next section, a number of processes for
revising epistemic states have been considered in the literature, and in
fact they all do satisfy these modified postulates.

Finally, even if we restrict attention to belief sets, we can consider
what happens if the underlying language  ${\cal L}$
is rich enough to talk about
how revision should be carried out.  For example, suppose ${\cal L}$
includes
conditional formulas, and we want to find some
ranking $\preceq$ for which the corresponding belief set is $K$.  Not
just any ranking $\preceq$ such that $\BEL(\preceq) = K$ will do here.
The beliefs in $K$ put some constraints on the ranking.  For
example, if $p \RCond q$ is in $K$ and $p \notin K$, then the minimal
$\preceq$-worlds satisfying $p$ must all satisfy $q$,
since after $p$ is learnt, $q$ is believed.
Once we restrict to rankings $\preceq_K$ that are consistent with $K$
then the AGM postulates are no longer sound.
This point has essentially been made before \cite{Boutilier92,Rott}.
However, it is worth stressing the sensitivity of the AGM postulates
to the underlying language and, more generally, to the choice of
epistemic state.
\section{Proposals for Iterated Revision}\label{otherproposals}

We now briefly review some of the previous proposals for iterated
belief change, and point out how the impact of the observations we have
been making on the approaches.  Most of these approaches start with the
AGM postulates, and augment them to get seemingly appropriate
restrictions on iterated revision.
This is not an exhaustive review of the literature on iterated belief
revision by any stretch of the imagination.  Rather, we have chosen a
few representative approaches that allow us to bring out our
methodological concerns.

\subsection{Boutilier's Approach}
As we said in the previous section, Boutilier
takes the agent's epistemic state to consist of
a ranking of possible worlds.
Boutilier \citeyear{Boutilier:Iterated} describes a
particular revision operator $\RevB$
on epistemic states.
This revision operator maps
a ranking $\preceq$ of possible worlds and an observation $\phi$ to a
revised ranking $\preceq \RevB \phi$ such that (a) $\preceq \RevB \phi$
satisfies the conditions of the
representation theorem described above, that is, the minimal worlds
in $\preceq \RevB \phi$ are
precisely the minimal $\phi$-worlds in $\preceq$,
and (b) in a precise
sense, $\preceq \RevB \phi$ is the result of making the minimal
number of changes to $\preceq$ required to guarantee that all the
minimal worlds in $\preceq \RevB \phi$ satisfy $\phi$.
Given a ranking $\preceq$ and a formula $\phi$, the
ranking $\preceq \RevB \phi$ is identical to $\preceq$ except that the
minimal $\phi$-worlds according to $\preceq$ have the minimal rank in
the revised ranking, while the relative ranks of all other worlds
remains unchanged.

Boutilier characterizes the properties of his approach
as follows.
Suppose
that, starting in some epistemic state, we revise by
$\phi_1, \ldots, \phi_n$.  Further suppose
$\phi_{i+1}$ is consistent with the beliefs
after revising by $\phi_1,\ldots,\phi_i$. Then the beliefs after
revising by $\phi_1, \ldots, \phi_n$ are precisely the
beliefs after observing $\phi_1\land\ldots\land\phi_n$.  (More
precisely, given any ranking $\preceq$,
the belief set associated with the ranking $\preceq \RevB
\phi_1 \RevB \ldots \RevB \phi_n$ is the same as that associated with
the ranking $\preceq \RevB (\phi_1 \land \ldots \land \phi_n)$.
Note, however, that  $\preceq \RevB \phi_1 \RevB
\ldots \RevB \phi_n \ne \preceq \RevB (\phi \land \ldots \land \phi_n)$
in general.)
Thus, as long
as the agent's new observations are not surprising, the agent's beliefs
are exactly the ones she would have had had she observed the
conjunction of all the observations.
This is an immediate consequence
of the AGM postulates, and thus holds for any approach
that attempts to extend the AGM postulates to iterated
revision.

\commentout{
Note,
however, that it is not the case that $\preceq \RevB \phi_1 \RevB
\ldots \RevB \phi_n = \preceq \RevB (\phi \land \ldots \land \phi_n)$;
these are different rankings that are associated with the same belief
set.  For example, suppose $\preceq_0$ makes all worlds minimal (so that
$\BEL(\preceq_0)$ consists of all tautologies).  If $p_1, \ldots, p_n$ are
primitive propositions, then $\preceq_0 \RevB p_1 \RevB \ldots \RevB
p_n$ results in a ranking where the minimal worlds are those
satisfying $p_1 \land \ldots \land p_n$, followed by worlds satisfying
$p_1 \land \ldots \land p_{n-1} \land \neg p_n$, followed by worlds
satisfying $p_1 \land \ldots \land p_{n-2} \land \neg p_{n-1}$, \ldots,
followed by worlds satisfying $p_1 \land \neg p_2$, and ending with the
worlds satisfying $\neg p_1$.  On the other hand, $\preceq_0 \RevB
(p_1 \land \ldots \land p_n)$ has only two
ranks; the minimal
rank consists of the worlds satisfying $p_1 \land \ldots \land p_n$,
and the other rank consists of all the other worlds.
}

What happens when the agent
observes a formula $\phi_{n+1}$ that is inconsistent with
her
current beliefs?
Boutilier shows that in this case the new observation
nullifies the impact of the all the observations starting with the most
recent one that is inconsistent with $\phi_{n+1}$. More precisely,
suppose $\phi_{i+1}$ is consistent with the belief after observing
$\phi_1, \ldots, \phi_i$ for
all
$i \le n$, but $\phi_{n+1}$ is inconsistent
with the beliefs after observing $\phi_1, \ldots, \phi_n$.  Let
$k$ be the maximal index
such that $\phi_{n+1}$ is consistent with the beliefs after learning
$\phi_1,\ldots,\phi_{k}$. The agent's beliefs after observing
$\phi_{n+1}$ are the same as her beliefs after observing
$\phi_1,\ldots,\phi_k,\phi_{n+1}$. Thus, the agent acts as though she
did not observe $\phi_{k+1}, \ldots, \phi_{n}$.
\commentout{
For example, it is easy
to see that the minimal worlds according to $\preceq_0 \RevB p_1 \RevB
\ldots \RevB p_n \RevB \neg p_j$ are those satisfying $p_0 \land
\ldots \land p_{j-1} \land \neg p_j$, with the rest of the order just
the same as that for $\preceq_0 \RevB p_1 \RevB \ldots \RevB p_n$:
the worlds satisfying $p_1 \land \ldots \land p_n$ come next in the
ranking, followed by those satisfying $p_1 \land \ldots \land p_{n-1}
\land \neg p_n$ (unless $j = n$), and so on.
}

Boutilier does not provide any argument for the reasonableness of this
ontology.  In fact, Boutilier's presentation (like almost all others
in the literature) is not in terms of an ontology at all; he presents
his approach as an attempt to minimize changes to the ranking.
While the intuition of minimizing changes to the ranking seems
reasonable at first, it becomes less reasonable when we realize its
ontological implications.  The following example, due to Darwiche and
Pearl \citeyear{Darwiche94}, emphasizes this point.  Suppose we
encounter a strange new animal and it appears to be a bird, so we
believe it is a bird.  On closer inspection, we see that it is red, so
we believe that it is a red bird.  However, an expert then informs us
that it is not a bird, but a mammal.
Applying Boutilier's revision operator, we
would no longer believe that the animal is red.  This does not seem so
reasonable.

One more point is worth observing:
As described by Boutilier \citeyear{Boutilier:Iterated},
his approach
does not allow revision by $\false$.
While we could, of course, modify the definition to handle $\false$, it
is more natural simply to disallow it.  This suggests
that, whatever ontology is used to justify Boutilier's approach, in that
ontology, revising by $\false$ should not make sense.

\subsection{Freund and Lehmann's Approach}
Freund and Lehmann \citeyear{FreundLehmann} stick close to the original
AGM approach. They work with belief sets, not more general epistemic
states.  However, they are interested in iterated revision.  They
consider the effect of adding just one more postulate to the basic
AGM postulates, namely
\begin{description}\denselist
\item[FL.] If $\neg \phi \in K$, then $K \Rev \phi = K_\bot \Rev \phi$,
\end{description}
where $K_\bot$ is the inconsistent belief set, which consists of all
formulas.

Suppose $\Rev$ satisfies R1--R8 and FL.
Just as with Boutilier's approach, if
$\phi_{i+1}$ is consistent with the beliefs after learning $\phi_1,
\ldots, \phi_i$ for $i \le n-1$, then
$K \Rev \phi_1 \Rev \ldots
\Rev \phi_n = K \Rev (\phi_1 \land \ldots \land \phi_n)$.  However, if
we then
observe $\phi_{n+1}$, and it is inconsistent with $K \Rev \phi_1 \land
\ldots \land \phi_n$, then $K \Rev \phi_1 \Rev \ldots \Rev \phi_{n+1} =
K_\bot \Rev \phi_{n+1}$.  That is, observing something inconsistent
causes us to retain none of our previous beliefs, but to start over from
scratch.  While the ontology here is quite simple to explain, as Freund
and Lehmann themselves admit, it is a rather severe form of belief
revision.  Darwiche and Pearl's red bird example applies to this
approach as well.

\subsection{Darwiche and Pearl's Approach}
Darwiche and Pearl \citeyear{Darwiche94}
suggest a set of postulates extending the AGM postulates, and
claim to provide a semantics that satisfies them.
Their intuition is that the
revision operator should retain
as much as possible certain parts of
the ordering among worlds in the ranking.  In particular,
if $w$ and $w'$ both satisfy $\phi$, then a
revision by $\phi$ should not change the relative rank of $w$ and
$w'$.  Similarly,
if both $w$ and $w'$ satisfy $\neg \phi$, then a revision should not
change their relative rank.
They describe four postulates that are meant to embody these
intuitions:
\begin{description}\denselist
\item[C1.] If $\phi \vdash \psi$, then $(K \Rev \psi) \Rev
\phi = K \Rev \phi$
\item[C2.] If $\phi \vdash \neg \psi$, then $(K \Rev \psi) \Rev
\phi = K \Rev \phi$
\item[C3.] If $\psi \in K \Rev \phi$, then $\psi \in (K \Rev \psi) \Rev
\phi$
\item[C4.] If $\neg \psi \notin K \Rev \phi$, then $\neg \psi \notin (K
\Rev \psi) \Rev \phi$
\end{description}

Freund and Lehmann \citeyear{FreundLehmann} point out that C2 is
inconsistent with the AGM postulates.  This observation seems
inconsistent with the fact that Darwiche and Pearl claim to
provide an example of a revision method that is consistent with their
postulates.
What is going on here?  It turns out that the issues raised earlier help
clarify the situation.

The semantics that Darwiche and Pearl use as an example is based on
a special case of Spohn's {\em ordinal conditional functions\/} (OCFs)
\citeyear{spohn:88}
called {\em $\kappa$-rankings} \cite{Goldszmidt92}.  A $\kappa$-ranking
associates
with each world either a natural number $n$ or $\infty$, with the
requirement that for at least one world $w_0$, we have $\kappa(w_0) =
0$. We can think
of $\kappa(w)$ as the rank of $w$, or as denoting how surprising it
would be to discover that $w$ is the actual world.
If $\kappa(w) = 0$, then world $w$ is unsurprising; if $\kappa(w) =
1$, then $w$ is somewhat surprising; if $\kappa(w) = 2$, then $w$ is
more surprising, and so on.  If $\kappa(w) = \infty$, then $w$ is
impossible.%
\footnote{Spohn allowed ranks to be
arbitrary ordinals, not just natural numbers, and did not allow a rank
of $\infty$, since, for philosophical reasons, he did not want to allow
a world to be considered impossible.  As we shall see, there are
technical advantages to introducing a rank of $\infty$.}
OCFs provide a way of ranking worlds that is closely related to, but
has a little more structure than the orderings considered by
Boutilier (as well as Grove and Katsuno and Mendelzon).
The extra structure makes it easier to define a notion of conditioning.

Given a formula $\phi$, let $\kappa(\phi) = \min\{\kappa(w): w \sat
\phi\}$; we define $\kappa(\false) = \infty$.   We say that {\em $\phi$
is believed with firmness $\alpha
\ge 0$ in OCF $\kappa$\/} if $\kappa(\phi) = 0$ and $\kappa(\neg \phi) =
\alpha$.  Thus, $\phi$ is
believed with firmness $\alpha$ if
$\phi$ is unsurprising and the least surprising world satisfying $\neg
\phi$ has rank $\alpha$.
We define $\BEL(\kappa)$ to
consist of all formulas that are believed
with firmness at least 1.

Spohn defined a notion of conditioning on OCFs.
Given an OCF $\kappa$, a formula $\phi$ such that $\kappa(\phi) <
\infty$, and $\alpha \ge 0$, $\kappa_{\phi,\alpha}$ is the unique OCF
satisfying the property desired by Darwiche and Pearl---namely,
if $w$ and $w'$ both satisfy $\phi$ or both satisfy
$\neg \phi$, then
revision by $\phi$ should not change the relative rank of $w$
and $w'$, that is,
$\kappa_{\phi,\alpha}(w) - \kappa_{\phi,\alpha}(w') =
\kappa(w) - \kappa(w')$---such that $\phi$
is believed with firmness $\alpha$ in
$\kappa_{\phi,\alpha}$.  It is defined as follows:
$$
\kappa_{\phi,\alpha}(w) = \left\{
\begin{array}{ll}
\kappa(w) - \kappa(\phi) &\mbox{if $w$ satisfies $\phi$}\\
\kappa(w) - \kappa(\neg \phi) + \alpha &\mbox{if $w$ satisfies $\neg
\phi$.}
\end{array}
\right.
$$
Notice that $\kappa_{\phi,\alpha}$ is defined only if $\kappa(\phi) <
\infty$, that is, if $\phi$ is considered possible.

Darwiche and Pearl defined the following revision function on OCFs:
$$\kappa \RevDP \phi = \left\{
\begin{array}{ll}
\kappa & \mbox{if
$\kappa(\neg\phi) \ge 1$
}\\
\kappa_{\phi,1} &\mbox{otherwise.}
\end{array}
\right.
$$
Thus, if $\phi$ is already believed with firmness at least 1 in
$\kappa$, then $\kappa$ is unaffected by a revision by $\phi$;
otherwise,
the effect of revision is to modify $\kappa$ by conditioning so that
$\phi$ ends up being believed with degree of firmness 1.
Intuitively, this means that if $\phi$ is not believed in $\kappa$, in
$\kappa\Rev \phi$ it is believed, but with the minimal degree of
firmness.

\commentout{
For example, if $\kappa_0$ assigns all
worlds a rank of 0
(so that $\BEL(\kappa_0) = \BEL(\preceq_0)$, and consists of all
tautologies), then $\kappa_0 \RevDP p_1 \RevDP \ldots \RevDP p_n$
assigns rank 0 to the worlds satisfying $p_1 \land \ldots \land p_n$,
rank 1 to the worlds satisfying $p_1 \land \ldots \land p_{n-1} \land
\neg p_n$, \ldots, rank $n-1$ to worlds satisfying $p_1 \land \neg p_2$,
and rank $n$ to worlds satisfying $\neg p_n$.  Thus, there is still some
preference for early information according to Darwiche and Pearl's
semantics for revision.
 When we next revise by
$\neg p_j$, the worlds satisfying $p_1 \land \ldots \land p_{j-1} \land
\neg p_j$ are assigned rank 0 (so these are the beliefs, just as with
natural revision), but the worlds satisfying $p_1 \land \ldots p_{j-2}
\land \neg p_{j-1} \land \neg p_j$ are assigned rank 1, whereas natural
revision would keep them at rank $j+1$.  This difference has an impact
on subsequent revisions.
}

It is not hard to show that if we take an agent's epistemic state to be
represented by an OCF, then Darwiche and Pearl's semantics satisfies all
the AGM
postulates modified to apply to epistemic states (that is, R1$'$--R8$'$
in Section~\ref{AGM}), except that revising by $\false$ is
disallowed, just as in Boutilier's approach, so that R5$'$ holds vacuously;
in addition, this semantics satisfies Darwiche and Pearl's
C1--C4, modified to apply to epistemic states.  For example,
C2 becomes
\begin{description}\denselist
\item[C2$'.$] If $\phi \vdash \neg \psi$, then $\BEL((E \Rev \psi)
\Rev \phi) = \BEL(E \Rev \phi)$.
\end{description}
Indeed, as Darwiche and Pearl observe, Boutilier's revision operator
also satisfies
C1$'$--C4$'$; however, it has properties that they view as undesirable.
Thus, Darwiche and Pearl's claim that their postulates are
consistent with AGM is correct, if we think at the level of general
epistemic states.  On the other hand, Freund and Lehmann are quite right
that
R1--R8 and C1--C4 are incompatible; indeed, as they point out, R1--R4
and C2 are incompatible.  The importance of making clear exactly whether
we are considering the postulates with respect to the OCF $\kappa$ or
the belief set $\BEL(\kappa)$ is particularly apparent here.%
\footnote{We note that in a recent version of their paper, Darwiche and
Pearl \citeyear{Darwiche94Full}
use a similar technique to deal with
the inconsistency of C2.}

The fact that Boutilier's revision operator also satisfies C1$'$--C4$'$
clearly shows that these postulates do not capture all of Darwiche and
Pearl's intuitions.  Their semantics embodies further assumptions.
Some of them seem {\em ad hoc}.  Why is it reasonable to believe
$\phi$ with a {\em minimal\/} degree of firmness after revising by
$\phi$?
Rather than
trying to come up with an improved collection of postulates
(which Darwiche and Pearl themselves suggest might be a difficult
task), it seems to us that a more promising
approach is to find an appropriate ontology.

\subsection{Lehmann's Revised Approach}
Finally, we consider Lehmann's ``revised'' approach to belief revision
\citeyear{Lehmann95}.
With each sequence $\sigma$ of observations, Lehmann associates a
belief set that we denote $\BEL(\sigma)$.
Intuitively, we can think of $\BEL(\sigma)$ as describing the agent's
beliefs after
making the sequence $\sigma$ of observations, starting from her initial
epistemic
state.
Lehmann allows all possible sequences of consistent formulas. Thus, he
assumes that the agent does not observe $\false$.
We view Lehmann's approach essentially as
taking the agent's epistemic state to be
the sequence of observations made, with the obvious revision operator
that concatenate a new observation to the current epistemic state. The
properties of belief change depend on the function $\BEL$.
Lehmann requires $\BEL$ to satisfy the following postulates (where
$\sigma$ and $\rho$ denote sequences of formulas, and $\cdot$ is the
concatenation operator):
\begin{description}\denselist
 \item[I1.] $\BEL(\sigma)$ is a consistent belief set
 \item[I2.] $\phi \in \BEL(\sigma\cdot\phi)$
 \item[I3.] If $\psi \in \BEL(\sigma\cdot\phi)$, then $\phi\rimp\psi
\in \BEL(\sigma)$
 \item[I4.] If $\phi \in \BEL(\sigma)$, then
$\BEL(\sigma\cdot\phi\cdot\rho) = \BEL(\sigma\cdot\rho)$
 \item[I5.] If $\psi \vdash \phi$, then
$\BEL(\sigma\cdot\phi\cdot\psi\cdot \rho) =  \BEL(\sigma\cdot\psi\cdot
\rho)$
 \item[I6.] If $\neg\psi \not\in \BEL(\sigma\cdot\phi)$, then
$\BEL(\sigma\cdot\phi\cdot\psi\cdot \rho) =
\BEL(\sigma\cdot\phi\cdot\phi\land\psi\cdot \rho)$
 \item[I7.] $\BEL(\sigma\cdot\neg\phi\cdot\phi) \subseteq
\Cl(\BEL(\sigma) \union \{ \phi \} )$
\end{description}
We refer the interested reader to \cite{Lehmann95} for the motivation
for these postulates. As Lehmann
argues, the spirit of the original AGM postulates is captured by these
postulates. Lehmann views
I5 and I7 as two main additions to the basic AGM postulates.  He
states that ``Since postulates I5 and I7 seem secure, \ie difficult to
reject, the postulates I1--I7 may probably be considered as a
reasonable formalization of the intuitions of AGM''
\cite[Section 5]{Lehmann95}.
Our view is that it is impossible to decide whether to accept or reject
postulates such as I5 or I7 (or, for that matter, any of the other
postulates) without an explicit ontology.  There may be ontologies for
which I5 and I7 are reasonable, and others for which they are not.
``Reasonableness'' is not an independently defined notion; it depends on
the ontology.
The ontology of the next section emphasizes this point.

\section{Taking Observations to be Knowledge}
\label{observationknowledge}

We now consider an ontology where observations are taken to be
knowledge.%
\footnote{We remark that Rott \citeyear[Section 6]{Rott91} outlines an
approach where obervations are treated as knowledge, but
does not provide an underlying ontology.
The high-level intuition behind his approach is similar to the one we
present here, although the details differ.}

As we said in the introduction, in this ontology, the agent has
some (closed) set of formulas that he {\em knows\/} to be true,
which is included in a larger set of formulas that he {\em believes\/}
to be true.
The belief set can be viewed as the result of applying some
nonmonotonic reasoning system grounded in the observations.
We can think of there being an ordering on the
strength of his beliefs, with the formulas known to be true%
---the observations and their consequences---%
having the greatest strength of belief.
Because observations are
taken to be knowledge, any formula observed is added to the
stock of knowledge (and must be consistent with what was
previously known).
In this ontology, it is
impossible to observe $\false$.  In fact, it is impossible to make any
inconsistent sequence of observations.  That is, if $\phi_1, \ldots,
\phi_n$ is observed, then $\phi_1 \land \ldots \land \phi_n$ must be
consistent (although it may not be consistent with the agent's original
beliefs).

In earlier work \cite{FrThesis,FrH2Full}, we presented one way of
formalizing this ontology, based on
the framework of Halpern and Fagin \citeyear{HF87} for modeling
multi-agent systems (see \cite{FHMV} for more details).
For modeling belief revision, we use this framework restricted to a
single agent.
\footnote{Although the multi-agent aspects of this framework does not
play a role in our analysis here, we note that it allows for a natural
extension of our results to multi-agent belief revision. This, however,
is beyond the scope of this paper.}
The key assumption in this framework
is that we can characterize the system by describing it in terms of a
{\em state\/} that changes over time.  Formally, we assume that at
each point in time, the agent is in some {\em local state}.
Intuitively, this local state encodes the information the agent has
observed thus far. There is also an {\em environment}, whose state
encodes relevant aspects of the system that are not part of the
agent's local state.  A {\em global state\/} is a tuple $(s_e, s_a)$
consisting of the environment state $s_e$ and the local state $s_a$ of
the agent. A {\em run\/} of the system is a function from time (which,
for ease of exposition, we assume ranges over the natural numbers) to
global states. Thus, if $r$ is a run, then $r(0), r(1), \ldots$ is a
sequence of global states that, roughly speaking, is a complete
description of what happens over time in one possible execution of the
system. We take a {\em system\/} to consist of a set of runs.
Intuitively, these runs describe all the possible behaviors of the
system, that is, all the possible sequences of events that could occur
in the system over time.

Given a system $\R$, we refer to a pair $(r,m)$ consisting of a run $r
\in \R$ and a time $m$ as a {\em point}. If $r(m) = (s_e, s_a)$, we
define $r_a(m) = s_a$ and $r_e(m) = s_e$.  We say two points $(r,m)$
and $(r',m')$ are {\em indistinguishable\/} to the agent, and write
$(r,m) \sim_a (r',m')$, if $r_a(m) = r'_a(m')$, \ie if the agent has
the same local state at both points.  Finally,
an {\em interpreted system\/} is a tuple $(\R,\pi)$,
consisting of a system $\R$ together with a mapping $\pi$ that
associates with each point a truth assignment to the primitive
propositions.

To capture the AGM framework, we consider a special class
of interpreted systems: We fix a propositional
language
${\cal L}$.  We assume
that the agent makes observations, which are characterized by formulas
in ${\cal L}$, and that her local state consists
of the sequence of observations that she has made.  We assume that the
environment's local state describes which formulas are actually true in
the world, so that it is a truth assignment to the formulas in ${\cal
L}$.  As observed by Katsuno and Mendelzon \citeyear{KM91}, the AGM
postulates assume that the world is {\em static}; to capture this, we
assume that the environment state does not change over time.  Formally,
we are interested in the unique interpreted system $(\R^{AGM},\pi)$ that
consists of all runs satisfying the following two assumptions for every
point $(r,m)$:
\begin{itemize}\denselist
\item The environment's state $r_e(m)$ is a truth assignment to the
formulas in ${\cal L}$ that agrees with $\pi$ at $(r,m)$ (that is,
 $\pi(r,m) = r_e(m)$), and $r_e(m) = r_e(0)$.
\item The agent's state $r_a(m)$ is a sequence of the form $\langle
\phi_1, \ldots,
\phi_m \rangle$, such that $\phi_1 \land \ldots \land \phi_m$ is
true according to the truth assignment $r_e(m)$ and $r_a(m-1) = \langle
\phi_1, \ldots, \phi_{m-1} \rangle$.
\end{itemize}
Notice that the form of the agent's state makes explicit an important
implicit assumption: that the agent remembers all her previous
observations.

In an interpreted system, we can talk about an agent's
knowledge: the agent knows $\phi$ at a point $(r,m)$ if $\phi$ holds
in all points $(r',m')$ such that $(r,m) \sim_a (r',m')$.
It is easy to see that, according to this definition, if $r_a(m) = \<
\phi_1, \ldots, \phi_m\>$, then the agent knows $\phi_1 \land \ldots
\land \phi_m$ at the point $(r,m)$: the agent's observations are known
to be true in this approach.
We are interested in talking about the agent's beliefs as well as her
knowledge.
To allow this, we added a
notion of {\em plausibility\/} to interpreted systems in
\cite{FrH1Full}.  We consider a variant of
this approach here, using OCFs, since it makes it easier to relate
our observations to Darwiche and Pearl's framework.

We assume that we start with an OCF $\kappa$ on runs such that
$\kappa(r) \ne \infty$ for any run $r$.  Intuitively, $\kappa$
represents our prior ranking on runs.  Initially, no runs is viewed as
impossible.
We then associate, with each point $(r,m)$, an OCF $\kappa^{(r,m)}$
on the runs.  We define $\kappa^{(r,m)}$ by induction on $m$.
We take $\kappa^{(r,0)} = \kappa$, and we take $\kappa^{(r,m+1)} =
\kappa^{(r,m)}_{\phi_{m+1},\infty}$, where $r_a(m+1) = \<\phi_1,\ldots,
\phi_{m+1}\>$.  Thus, $\kappa^{(r,m+1)}$ is the result of conditioning
$\kappa^{(r,m)}$ on the last observation the agent made, giving it
degree of firmness $\infty$.
Thus, the agent is treating
the observations as knowledge in a manner compatible with the semantics
for knowledge in
interpreted systems.
Moreover, since observations are known, they are also believed.

Note that we could have described the same belief change process by
considering an OCF on formulas, rather than runs.  However, we feel that
using runs and systems provides a better model of what is going on, and
gives a more explicit ontology.  We can then {\em derive\/} an OCF on
formulas from the OCF on runs that we consider.

\commentout{
As we show in \cite{FrThesis,FrH2Full}, this framework
satisfies the AGM postulates
R1$'$--R8$'$, interpreted on epistemic states.  (Here we take the
agent's epistemic state at the point $(r,m)$ to consist of $r_a(m)$
together with $\kappa^{(r,m)}$.)  Moreover,
it is easy to verify that
the framework also satisfies
Darwiche and Pearl's postulates (appropriately modified to apply to
epistemic states), except that the contentious C2 is now vacuous, since
it is illegal to revise by $\psi$ and then $\phi$ if $\phi \vdash \neg
\psi$.
}
As we show in \cite{FrThesis,FrH2Full}, this framework
satisfies the AGM postulates R1$'$--R8$'$, interpreted on epistemic
states.  (Here we take the agent's epistemic state at the point
$(r,m)$ to consist of $r_a(m)$ together with $\kappa^{(r,m)}$.)
Moreover, we show this framework satisfies
an additional postulate,
which we call R9$'$:
\begin{description}\denselist
 \item[(R9$'$)] If $\not\vdash_\L \neg(\phi \land \psi)$ then $\BEL(E \Rev
 \phi \Rev\psi) = \BEL(E\Rev \phi\land\psi)$.
\end{description}
This postulate captures the intuition that observations are taken to
be knowledge, and thus observing $\phi$ and then $\psi$ is equivalent
to observing $\phi \land \psi$.
In fact, we show there that R1$'$--R9$'$  characterize, in a precise
sense, revision in this framework.

It is easy to verify that the framework also satisfies Darwiche and
Pearl's postulates (appropriately modified to apply to epistemic
states), except that the contentious C2 is now vacuous, since it is
illegal to revise by $\psi$ and then by $\phi$ if $\phi \vdash \neg
\psi$.

How does this framework compare to Lehmann's?  Like Lehmann's, there is
an explicit attempt to associate beliefs with a sequence of revisions.
However, we have restricted the sequence of revisions, since we are
treating observations as knowledge.  It is easy to see that I1--I3 and
I5--I7 hold in our framework.  However, since we have restricted the
sequence of observations allowed, some of these postulates are much
weaker in our framework than in Lehmann's.  In particular, I7 is satisfied
vacuously, since we do not allow
a sequence of the form $\sigma\cdot\neg\phi\cdot\phi$.
On the other hand, I4 is not satisfied in our framework.  Our discussion
in the introduction suggests a counterexample.  Suppose that initially,
$\kappa(p \land q) = 0$, $\kappa(\neg p \land q) = 1$, $\kappa(p \land
\neg q) = 2$, and $\kappa(\neg p \land \neg q) = 3$.  Thus, initially
the agent believes both $p$ and $q$, but believes $p$ with firmness 1
and
$q$ with firmness 2.  If the agent then observes $\neg p \lor \neg
q$, he will then believe $q$ but not $p$.  On the other hand, suppose
the agent first observes $p$.  He still
believes both $p$ and $q$, of course, but now $p$ is believed with
firmness $\infty$.  That means if he then observes $\neg p \lor \neg q$,
he will believe $p$, but not $q$, violating I4.
However, a weaker variant of I4 does hold in our
system: if the agent {\em knows\/} $\phi$, then observing $\phi$ will
not change her future beliefs.

Did we really need all the machinery of runs and systems here?
While we could no doubt get away without it, we believe that modeling
time and the agent's state explicitly adds a great deal.  In particular,
by including time, we capture the belief revision process
within the model.  By including the agent's state and the environment,
we can capture various assumptions about the
agent's observational capabilities (this is especially relevant once
we allow inaccurate observations) and generalize to allow multiple
agents.  See \cite{BFH1,FrH2Full,FrH1Full}
for further discussion and demonstration of the advantages of this
framework.

\section{Discussion}\label{conclusion}
The goal of this paper was to highlight what we see as some
methodological problems in much of the literature on belief revision.
There has been (in our opinion) too much attention paid to postulates,
and not enough to the underlying ontology.
An ontology must make clear what the agent's epistemic state is, what
types of observations the agent can make, the status of observations,
and how the agent goes about revising the epistemic state.

We have (deliberately) not been very precise about what counts as
an ontology, and clearly there are different levels of detail
that one can provide.  Although some papers on belief revision
have attempted to provide something in the way of an ontology,
the ontology has typically been insufficiently detailed to verify
the reasonableness of the postulates.  For example,
G\"ardenfors \citeyear{Gardenfors1}%
---whose ontology is far better developed than most---%
takes belief sets to consist
of formulas that are accepted, and revision to be by formulas
that are accepted.  As we have seen, unless belief sets or the
revision operator contain additional information (such as
epistemic importance or strengths of beliefs) this ontology will
violate R1.  On the other hand, unless we  are careful about how
we add such information, we may well violate some other axioms
(such as R3 or R5).  In any case, it is the job of the ontology to
make completely clear such issues as
whether observations are believed to
be true or known to be true, and if they are believed, what the strength
of belief is.  This issue is particularly important if we have epistemic
states like rankings that are richer than belief sets.  If observations
are believed, but not necessarily known, to be true, then it is not
clear how to go about revising such a richer epistemic state.  With
what degree of firmness should the new belief be held?  No particular
answer seems to us that well motivated.  It may be appropriate
for the user to attach degrees of firmness to observations, as was done
in \cite{GoldszmidtThesis,Williams94,Wobcke95}  (following the lead of
Spohn \citeyear{spohn:88});
we can even generalize to allowing uncertain
observations \cite{DP92}.

\commentout{
We are currently working on finding alternative ontologies where
observations are not taken to be knowledge.  One such an ontology,
motivated by ideas from stochastic processes, is suggested in
\cite{FrThesis} for   the case of update and can be easily adapted for
revision as well. (Boutilier~\cite{Boutilier95Full} describes a
similar ontology for update.)
} %
We are currently interested in finding alternative ontologies where
observations are not taken to be knowledge.  One such ontology,
which is motivated by ideas from stochastic processes, is described in
\cite{BFH1}.
This ontology explicitly models how ``noisy''
observations come about.
Roughly speaking, we assign a prior
plausibility to observing a formula $\phi$ in a world $w$. When we
observe $\phi$ we update our
plausibility measure on
worlds by considering the
plausibility of observing $\phi$ in each of these worlds. Of course, we
can then consider various assumptions
on this prior plausibility.
For example, we might say that
we believe observations to be true.  That is, our prior
plausibility of observing $\phi$ in worlds where it is true is higher
than the plausibility of observing $\phi$ in worlds where it false. We
note that this assumption does {\em not\/} imply that $\phi \in \BEL(K
\Rev \phi)$, since we have to combine the plausibility of observing
$\phi$ with
the plausibility of $\phi$ before the observation was made.  If $\phi$
is considered implausible in $K$, then the single observation of
$\phi$ might not suffice to make $\phi$ more plausible than
$\neg\phi$. (For example, the medieval scientist probably would not
change her beliefs about the speed of falling objects after a single
experiment.)
In this paper we have focused on belief revision. However,
the need to clarify the underlying ontology goes far beyond belief
revision.  Much the same comments can be made for all the work
on nonmonotonic logic as well (this point is essentially made in
\cite{Hal10}).  Not surprisingly, our
critique applies to other approaches belief change as well, and in
particular to Katsuno and Mendelzon's {\em belief update\/}
\citeyear{KM91}. Although the motivation described by Katsuno and
Mendelzon is different than that of revision, the discussion of update
is stated in terms of postulates about
belief sets. Thus, for
example, the distinction between the agent's belief set and
epistemic state arises in update as well: update is defined as
function from (belief sets $\cross$ formulas) to belief sets.
However, the agent's belief set does not uniquely determine the outcome
of update. This is demonstrated,
for example, by Katsuno and Mendelzon's semantic characterization that requires
the agent to have a ternary relation on possible worlds such that $w_1
<_w w_2$ holds if the agent considers $w_1$ to be ``closer'' to $w$
than $w_2$.  Other issues we raise here also apply to update for
similar reasons.

The ontology we propose in Section~\ref{observationknowledge}, where
observations are treated as knowledge, can be applied to update as
well.  Moreover, the assumption that observations are true is less
problematic for iterated update: since update does not assume that
propositions are static, we can consider runs where $\phi$ is true at
one time point, and false at the next one.  Thus, assuming that
observations are known to be true does not rule out sequences of
observations of the form $\phi, \neg\phi, \ldots$. In fact, all
sequences of consistent observations are allowed. We refer the reader
to \cite{FrThesis,FrH2Full} for more details.

It seems to us that many of the intuitions that researchers in the
area have are motivated by thinking in terms of observations as known,
even if this is not always reflected in the postulates considered.  We
have examined carefully one particular instantiation of this ontology,
that of treating observations as knowledge.
We have shown that, in this ontology,
some postulates that seem reasonable, such as Lehmann's I4, do not hold.
We do not mean to suggest that I4 is ``wrong'' (whatever that might
mean in this context).   Rather, it shows that we cannot blithely accept
postulates without making the underlying ontology clear.
We would encourage the investigation of other ontologies for belief
change.

\subsubsection*{Acknowledgments}
The authors are grateful to
Craig Boutilier, Adnan Darwiche,
Adam Grove,
Daniel Lehmann, and an anonymous reviewer for comments on
the paper and useful discussions relating to this work.

\bibliographystyle{chicago}
\bibliography{z,refs,conf-long}
\end{document}